\documentclass[10pt,twocolumn,letterpaper]{article}

\usepackage{cvpr}              

\usepackage[dvipsnames]{xcolor}

\usepackage{siunitx}
\definecolor{cvprblue}{rgb}{0.21,0.49,0.74}
\usepackage[pagebackref,breaklinks,colorlinks,citecolor=cvprblue]{hyperref}
\usepackage{multirow}
\usepackage{graphicx}
\usepackage{subcaption}
\usepackage{fontawesome}

\def\paperID{4158} 
\def\confName{CVPR}
\def\confYear{2024}

\title{View-decoupled Transformer for Person Re-identification \\ under Aerial-ground Camera Network}

\author{Quan Zhang$^{1,5}$ \and Lei Wang$^{1}$ \and Vishal M. Patel$^{5}$ \and Xiaohua Xie$^{1,2,3,4}$ \and Jianhuang Lai$^{1,2,3,4}$\thanks{Corresponding Author.} \and
$^{1}$School of Computer Science
and Engineering, Sun Yat-Sen University, China
\\
$^{2}$Pazhou Lab (HuangPu), Guangdong, China
\\
$^{3}$Guangdong Province Key Laboratory of Information Security Technology, Guangzhou, China
\\
$^{4}$Key Laboratory of Machine Intelligence
and Advanced Computing, Ministry of Education,
China
\\
$^{5}$Department of Electrical and Computer Engineering, Johns Hopkins University, USA
\\
{\tt\small \{zhangq48, wanglei75\}@mail2.sysu.edu.cn, 
\{stsljh, xiexiaoh6\}@mail.sysu.edu.cn,
}
\\
{\tt\small vpatel36@jhu.edu
}
}

\begin{document}
\maketitle
\begin{abstract}
Existing person re-identification methods have achieved remarkable advances in appearance-based identity association across homogeneous cameras, such as ground-ground matching. However, as a more practical scenario, aerial-ground person re-identification (AGPReID) among heterogeneous cameras has received minimal attention. To alleviate the disruption of discriminative identity representation by dramatic view discrepancy as the most significant challenge in AGPReID, the view-decoupled transformer (VDT) is proposed as a simple yet effective framework. Two major components are designed in VDT to decouple view-related and view-unrelated features, namely hierarchical subtractive separation and orthogonal loss, where the former separates these two features inside the VDT, and the latter constrains these two to be independent. In addition, we contribute a large-scale AGPReID dataset called CARGO, consisting of five/eight aerial/ground cameras, 5,000 identities, and 108,563 images. Experiments on two datasets show that VDT is a feasible and effective solution for AGPReID, surpassing the previous method on mAP/Rank1 by up to 5.0\%/2.7\% on CARGO and 3.7\%/5.2\% on AG-ReID, keeping the same magnitude of computational complexity. Our project is available at \url{https://github.com/LinlyAC/VDT-AGPReID}.
\end{abstract}
    
\section{Introduction}
\label{sec:intro}
\begin{figure}[t]
	\centering
	\includegraphics[width=0.9\linewidth]{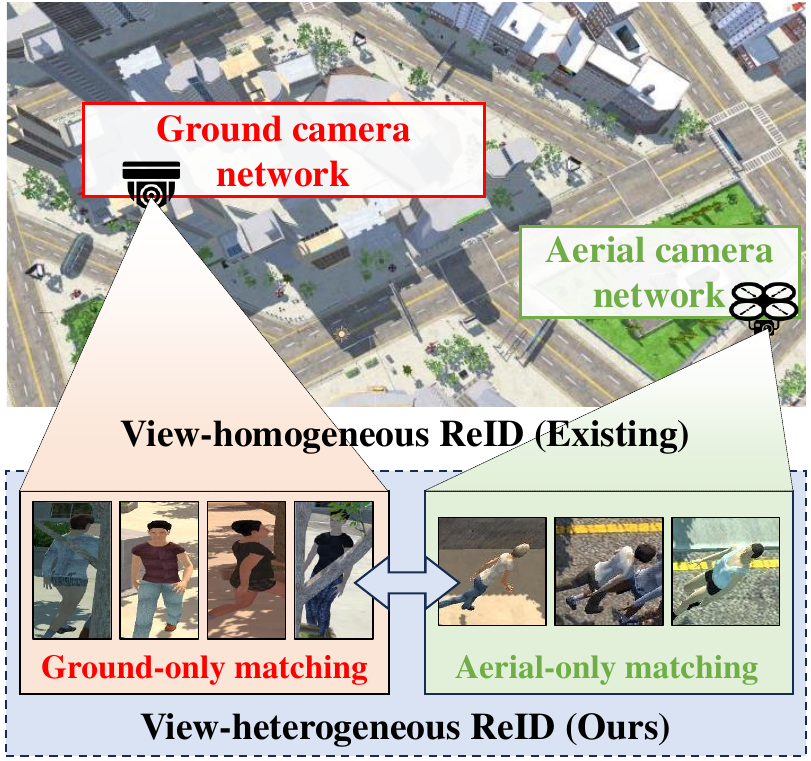}
	\caption{View-homogeneous vs. view-heterogeneous ReID, where the former focuses on ground-only or aerial-only camera networks, and the latter considers the aerial-ground mixed camera network. Thus, view-heterogeneous ReID considers aerial-aerial, ground-ground, and aerial-ground matching, which is more challenging and practical than the existing view-homogeneous ReID.}
	\label{fig1}
\end{figure}

Person re-identification (ReID) aims to associate interested person images based on cross-camera identity similarity, which plays a significant and positive contribution to the security and safety of society and citizens \cite{reid5, WePerson, sot, SSRG, City1M}. Traditional ReID has achieved remarkable progress in the deep learning era \cite{snn1, seg2, gan1, aerial2, gan2, snn2, gan3, seg1, PersonX, UMSOT,attn1}, but it is far from real-world scenarios for the following reasons. (1) \textit{Most existing ReID datasets are collected from homogeneous camera networks}, which comprises of the same type cameras, such as ground-only \cite{market} or aerial-only \cite{prai} camera networks. However, as can be seen from \cref{fig1}, real-world surveillance systems tend to be deployed as heterogeneous camera networks comprising aerial and ground cameras instead of a single type alone. Ground cameras cover well-developed areas (city centers), and aerial cameras cover poorly-developed areas (suburbs) due to their broad view range. The complementary nature of the two types would maximize the effectiveness of ReID. (2) \textit{Existing ReID methods mainly consider homogeneous matching (ground-ground \cite{cpn_tip, alder, AGW} and aerial-aerial \cite{AG-ReID, aerial1, aerial2}) }, which is ineffective in dealing with the dramatic view discrepancy among heterogeneous matching (aerial-ground). Despite its significance, related research is extremely scarce.

In this paper, we consider the novel and practical view-heterogeneous ReID problem, specifically ReID under the aerial-ground camera network (AGPReID). We propose a view-decoupled transformer (VDT) to specifically tackle the dramatic view discrepancy, which serves as a significant challenge within AGPReID to hinder homogeneous and heterogeneous matchings. Motivated by the fact that view-related features are useless for discriminative identity representations, VDT aims to decouple view-related and view-unrelated components, further facilitating discriminative identity learning from the remaining view-unrelated features. There are two key parts within VDT to achieve view decoupling, namely \textbf{(a) hierarchically subtractive separation} and \textbf{(b) orthogonal loss}. Specifically, after tokenizing the input image to a series of patch tokens, VDT appends two extra tokens (the meta token and view token) to feed them into a transformer stacked by multiple VDT blocks. The meta token captures global representation in the image, and the view token aims to extract view-related features. In each VDT block, (a) is achieved by the subtractive values of the meta token and view token after self-attention operations as the updated meta token for the next block. This means that VDT hierarchically separates view-related features from global features to facilitate identity learning from remaining view-unrelated features. When the meta and view tokens are obtained after the last block, (b) is designed to constrain the identity features to be independent of the view features, ultimately achieving the orthogonal decoupling of view-related and view-unrelated components. Meanwhile, the meta and view tokens will be supervised by identity and view labels, respectively. 

Considering the scarcity \cite{AG-ReID} and privacy \cite{UnrealPerson, City1M} of the AGPReID datasets, we contribute a large-scale \textbf{C}ivic \textbf{A}e\textbf{R}ial-\textbf{G}r\textbf{O}und (CARGO) dataset, which is collected from a synthetic city scenario including five aerial and eight ground cameras. CARGO has totally collected 5,000 identities and 108,563 images, which contain not only significant view discrepancy but also plenty of variations of resolution, illumination, occlusion, \etc. Regarding evaluation, CARGO considers multiple patterns (aerial-aerial, ground-ground, aerial-ground matchings), which could serve as a comprehensive benchmark. Contributions of this paper can be summarized as follows:
\begin{itemize}
	\item We focus on ReID under the aerial-ground camera network (AGPReID) scenario and propose a view-decoupled transformer (VDT) specifically for the dramatic view discrepancy within AGPReID, which achieves view-related and view-unrelated features decoupling by hierarchical subtractive separation and orthogonal loss. 
	\item We contribute a large-scale synthetic dataset, called CARGO, to advance the AGPReID as a benchmark, which contains five/eight aerial/ground cameras, 5,000 identities, 108,563 images, and complex challenges.
	\item Experiments demonstrate the superiority of VDT on two datasets, which shows that VDT surpasses the previous method on mAP/Rank1 by up to 5.0\%/2.7\% on CARGO and 3.7\%/5.2\% on AG-ReID, respectively, maintaining the same magnitude of computational complexity.
\end{itemize}

\section{Related Work}
\begin{figure*}[t]
	\centering
	\includegraphics[width=0.95\linewidth]{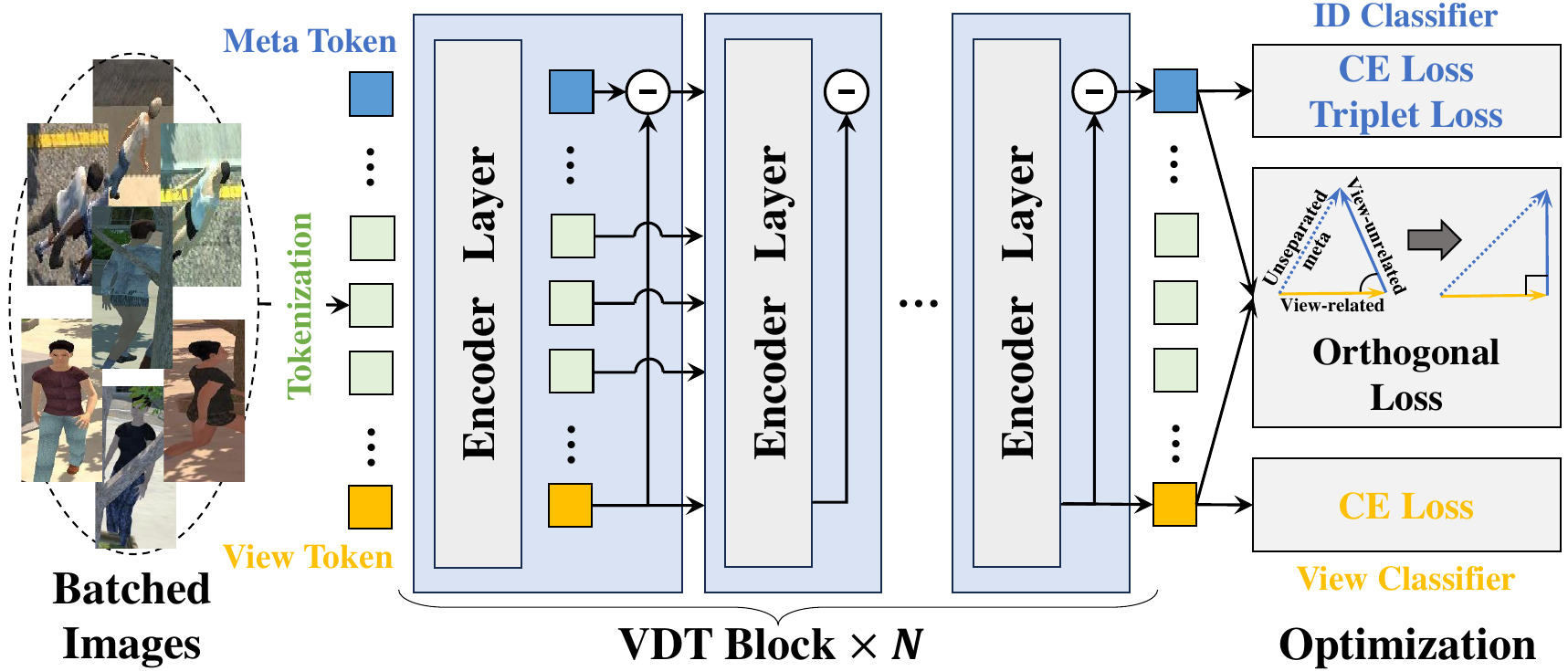}
	\caption{Illustration of the proposed VDT framework, which consists of $N$ VDT blocks and three-part loss functions. Meta and view tokens capture global and view-related features in images, respectively. Each VDT block (light blue module) consists of a standard self-attention encoder layer and an inner feature subtraction operation, achieving layer-by-layer decoupling of view-related and view-unrelated features. Orthogonal loss constrains the above two features to be further independent.}
	\label{fig:vdt}
\end{figure*}
\subsection{View-homogeneous ReID}\label{sec2.1} View-homogeneous ReID deals with images from ground-only or aerial-only camera networks, where the ground-only camera network receives the broadest attention in the ReID task because it is the most common use case. Many ground-only datasets have been contributed in the literature, such as Market1501 \cite{market}, MSMT17 \cite{msmt17}, \etc. Accordingly, lots of methods have been proposed to bring rapid development to ReID, including handcrafted feature-based methods \cite{histlbp_eccv, lomo_cvpr}, CNN-based methods \cite{AGW, PCB, alder, cpn_tip}, and transformer-based methods \cite{TransReID}. Regarding aerial-only research, only some pioneering efforts have contributed substantial datasets and methods \cite{prai, uavhuman, RotTrans} because it has not received as much attention as ground-only ReID.

However, it is difficult to directly transfer the above methods to the view-heterogeneous ReID because these methods ignore the significant view differences between aerial and ground cameras, leading to weak performances.

\subsection{View-heterogeneous ReID} 
Compared to the abundant research in \cref{sec2.1}, studies on view-heterogeneous ReID are almost nonexistent. In this paper, heterogeneous views refer to AGPReID, and the fundamental challenge is the dramatic view discrepancy. To the best of our knowledge, one recent work \cite{AG-ReID} has formally attempted to address this task. They collected an outdoor scene dataset, AG-ReID, with both identity and attribute labeling of pedestrians. Moreover, they proposed an explainable model by attribute information to guide model training.

While pioneering and critical to AGPReID, this work is still imperfect for the following reasons. (1) \textit{Dataset.} Compared to \cite{AG-ReID}, our CARGO dataset comprehensively focuses on multiple matching patterns that may appear in heterogeneous views rather than a single aerial-ground matching in AG-ReID. (2) \textit{Method.} Compared to explainable methods in \cite{AG-ReID}, our VDT relies on a lesser amount of priori labeling and thus has a stronger generalization. The experiments also show that our VDT achieves stronger performances.

\subsection{Synthetic ReID Dataset}
Data privacy is an inevitable issue for the ReID datasets, and synthetic data provides a feasible solution because it not only achieves privacy protection for real-world pedestrians but also has lower collection costs. Many synthetic ReID datasets have been published in the literature, such as PersonX \cite{PersonX}, UnrealPerson \cite{UnrealPerson}, WePerson \cite{WePerson}, \etc.

However, existing synthetic datasets mainly focus on simulating pedestrian data under view-homogeneous camera networks, and the simulation of aerial-ground views is poorly studied. To this end, we contribute the CARGO dataset as the first large-scale synthetic dataset for AGPReID.

\section{Method}
\subsection{Formulation and Overview}
\textbf{Formulation.} An AGPReID dataset $\mathcal{D}=\left\{ \left( x_i,y_i,v_i \right) \right\} _{i=1}^{|\mathcal{D}|}$ consists of a training set $\mathcal{D}^{tr}$ and a test set $\mathcal{D}^{te}$, where $x_i$ represents the $i$-th person image, and $y_i$ and $v_i$ stand for the corresponding identity and view labels, respectively. Note that $v_i \in \left\{v^a, v^g\right\}$ is readily acquired by the known camera labels in $\mathcal{D}$ to easily distinguish whether $x_i$ belongs to an aerial $v^a$ or ground $v^g$ view. Significant view discrepancy between $v^a$ and $v^g$ leads to a view-biased feature space, which behaves as low intra-identity and high inter-identity similarity. Therefore, the goal of AGPReID methods is to design a model $\mathcal{F}(\cdot; \theta_\mathcal{F})$ with learnable parameter $\theta_\mathcal{F}$ against the view bias, which can be written as:
\begin{equation}
	\underset{\theta}{\min}\sum_i{\left[ \begin{array}{c}
			\left\| \mathcal{F}\left( x_i;\theta_\mathcal{F} \right) -\mathcal{F}\left( x_{i}^{+};\theta_\mathcal{F} \right) \right\| _{2}^{2}-\\
			\left\| \mathcal{F}\left( x_i;\theta_\mathcal{F} \right) -\mathcal{F}\left( x_{i}^{-};\theta_\mathcal{F} \right) \right\| _{2}^{2}\\
		\end{array} \right]}, \label{goal}
\end{equation}
where $\|\cdot\|_2$ denotes the L2 distance, $x_{i}^{+}/x_{i}^{-}$ denotes the person image that has the same/different identity with $x_i$.

\noindent\textbf{Overview.} As shown in \cref{fig:vdt}, we design a simple yet effective framework called view-decoupled transformer (VDT), to tackle the view discrepancy challenge in AGPReID. For a batched data $B$ that contains both $v^a$ and $v^g$ view, we tokenize them as a series of image tokens (green squares), then append a meta and view token (blue and yellow squares) to them, which serve as input to VDT. The VDT network consists of $N$ blocks, where each block first performs the standard self-attention encoding and then performs a subtraction operation between meta and view tokens to explicitly separate the view-related feature from the global feature. Finally, the identity and view classifier will supervise the updated meta and view tokens outputted from VDT. Besides, an orthogonal loss is proposed that makes these two tokens irrelevant, thus achieving the complete decoupling of view-related and view-unrelated features.

\begin{table*}[t]
	\centering
	\caption{Statistical comparisons of existing datasets, including view-homogeneous (ground or aerial) and view-heterogeneous (ground and aerial) ReID. ``$\star$'' denotes City1M is used for group ReID, where its images correspond to the number of group images.}
	\label{datasets}
	\renewcommand{\arraystretch}{1.1}
	\resizebox{0.8\textwidth}{!}{%
		\begin{tabular}{|c|c|c|c|c|c|c|}
			\hline
			Dataset & View & Data &\#PersonID & \#Camera & \#Image & \#Height \\ \hline
			Market1501 \cite{market} & Ground & Real & 1,501 & 6 & 32,668 & $<10m$ \\
			PersonX \cite{PersonX} & Ground & Synthetic & 1,266 & 6 & 273,456 & - \\
			RandPerson \cite{RandPerson} & Ground & Synthetic & 8,000 & 19 & 228,655 & - \\
			UnrealPerson \cite{UnrealPerson} & Ground & Synthetic & 3,000 & 34 & 120,000 & - \\
			WePerson \cite{WePerson} & Ground & Synthetic & 1,500 & 40 & 4,000,000 & - \\
			ClonedPerson \cite{ClonedPerson} & Ground & Synthetic & 5,621 & 24 & 887,766 & - \\
			City1M$^\star$ \cite{City1M} & Ground & Synthetic & 45,000 & 8 & 1,840,000 & - \\ \hline
			PRAI1581 \cite{prai} & Aerial & Real & 1,581 & 2 & 39,461 & $20\sim60m$ \\
			UAVHuman \cite{uavhuman} & Aerial & Real & 1,144 & 1 & 41,290 & $2\sim8m$ \\ \hline
			AG-ReID \cite{AG-ReID} & Aerial-Ground & Real & 388 & 2 (1A+1G) & 21,893 & $15\sim45m$ \\
			\textbf{CARGO (Ours)} & \textbf{Aerial-Ground} & \textbf{Synthetic} & \textbf{5,000} & \textbf{13 (5A+8G)} & \textbf{108,563} & $5\sim75m$ \\ \hline
		\end{tabular}%
	}
\end{table*}

\subsection{View-decoupled Transformer} \label{sec3.2}
The proposed VDT framework is based on Vit-Base \cite{Vit}. For each image $x_i$ in the batched data $B$, VDT first divides $x_i$ evenly and non-overlappingly into $M$ image patches, which are further tokenized into $M$ patch embeddings by a 1$\times$1 convolution and can be denoted as $\left[t_{p,1}; t_{p,2}; \cdots; t_{p, M}\right]$.
After that, we append two additional learnable tokens, meta token $t_m$ and view token $t_v$, in the patch tokens, where $t_m$ aims to capture the global image representation, $t_v$ focuses only on the view-related feature. Next, we serialize all the tokens by assigning them the corresponding positional embedding, which can be written as:
\begin{equation}
	\mathcal{S}(x_i) \triangleq [t_m; t_{p,1}; t_{p,2}; \cdots; t_{p,M}; t_v] + \mathbf{E}_{pos},
\end{equation}
where $\mathcal{S}(\cdot)$ is defined as the serialization operation for $x_i$, $t_m, t_{p,i}, t_v \in \mathbb{R}^{d}$, and $\mathbf{E}_{pos} \in \mathbb{R}^{(M+2)\times d}$.

As shown in \cref{fig:vdt}, VDT $\mathcal{F}(\cdot; \theta_\mathcal{F})$ is stacked with $N$ VDT blocks $\mathcal{F}_j(\cdot; \theta_\mathcal{F}^j), 1\leq j\leq N$. Each $\mathcal{F}_j(\cdot; \theta_\mathcal{F}^j)$ contains two operations, self-attention encoding and subtractive separation, which are described as:
\begin{equation}
	[t_m^{(j+1)}(x_i); \!\cdots\!; t_v^{(j+1)}(x_i)] \!=\! \theta_\mathcal{F}^{(j)}([t_m^{(j)}(x_i); \!\cdots\!; t_v^{(j)}(x_i)]), \label{attn}
\end{equation}
\begin{equation}
	t_{m}^{(j+1)}(x_i) \leftarrow t_{m}^{(j+1)}(x_i)-t_{v}^{(j+1)}(x_i), \label{innersub}
\end{equation}
where $\theta_\mathcal{F}^{(j)}$ represents the self-attention  parameters in $\mathcal{F}_j$. If $j=1$, the input of  $\mathcal{F}_1$ becomes $\mathcal{S}(x_i)$. Note that $t_m^{(j+1)}(x_i)$ and $t_v^{(j+1)}(x_i)$ in \cref{attn} represent the updated meta and view token after the self-attention operation in $\mathcal{F}_j$ with patch tokens of $x_i$.  \cref{innersub} explicitly guides that $t_{m}^{(j+1)}$ and $t_{v}^{(j+1)}$ are distinguishable after \cref{attn}, where $t_{v}^{(j+1)}$ captures more view-related features and $t_{m}^{(j+1)}$ can captured more view-unrelated features by removing  $t_{v}^{(j+1)}$ from the global (meta) features. \cref{innersub} facilitates the learning of identity features via $t_{m}^{(j+1)}$, alleviating the disturbing of view bias. VDT decouples the view-related and -unrelated features layer by layer until the updated $t_{m}^{(N+1)}$ and  $t_{v}^{(N+1)}$ are obtained, which serves as the well-decoupled identity and view features, respectively. During inference, we only use $t_{m}^{(N+1)}$ as the whole retrieval evidence.

In addition, the proposed VDT maintains the same computational complexity as baseline (ViT). Given that the input to ViT has $T$ tokens and the dimensions of the tokens are $d$, then the computational complexity of ViT is $
\mathcal{O}\!\left( 4Td^2+2T^2d \right)$ \cite{swinT}. Our VDT adds an additional $t_v$ token based on the baseline and adds an \cref{innersub} operation in each VDT block. Therefore, the total computational complexity can be written as:
\begin{equation}
	\renewcommand\arraystretch{1.8}
	\begin{array}{l}
		\mathcal{O}\!\left( 4\left( T+1 \right) d^2+2\left( T+1 \right) ^2d+N \right)\\
		\Leftrightarrow \mathcal{O} \!\left( \underset{baseline\,\,complexity}{\underbrace{4Td^2+2T^2d}}+\underset{constant\,\,complexity}{\underbrace{4d^2+2d+N}} \right)\\
		\Leftrightarrow \mathcal{O} \!\left( 4Td^2+2T^2d \right),
	\end{array} \label{complex}
\end{equation}
where $N$ counts all \cref{innersub} in $N$ VDT blocks. \cref{complex} shows that the extra operations VDT introduces depend on only the network hyperparameters, which belong to constant complexity and can be omitted. Therefore,  VDT maintains the same magnitude of complexity as that of the baseline.

\begin{figure*}[t]
	\centering
	\begin{subfigure}{0.202\linewidth}
		\centering
		\includegraphics[width=\linewidth]{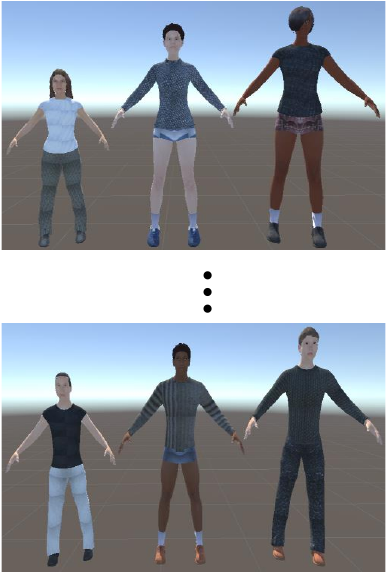}
		\caption{3D pedestrian models.}
		\label{fig3a}
	\end{subfigure}
	\begin{subfigure}{0.55\linewidth}
		\centering
		\includegraphics[width=\linewidth]{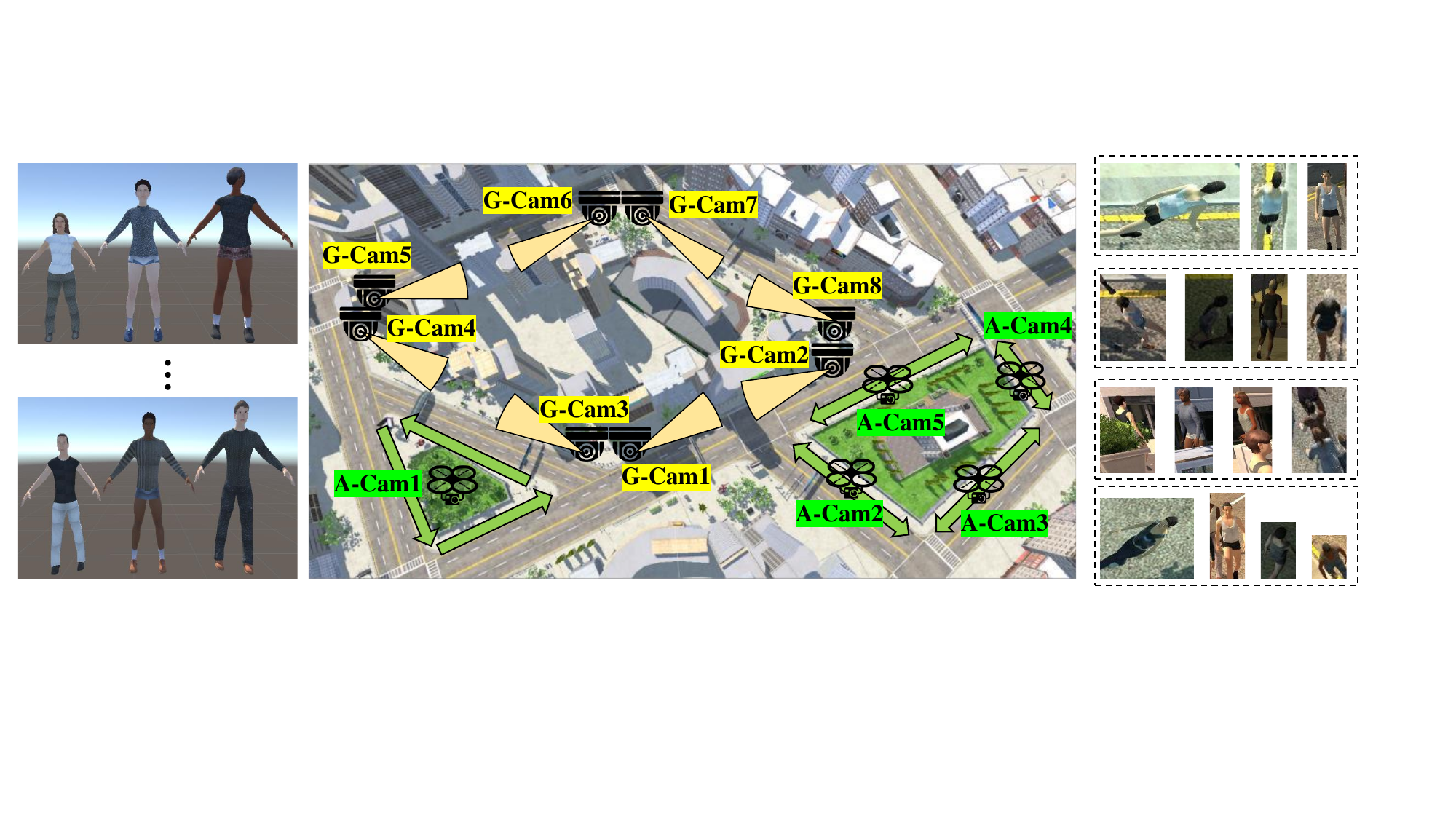}
		\caption{Aerial-ground camera network deployment.}
		\label{fig3b}
	\end{subfigure}
	\begin{subfigure}{0.188\linewidth}
		\centering
		\includegraphics[width=\linewidth]{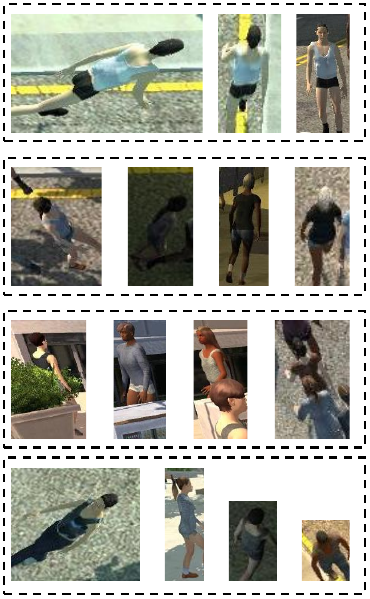}
		\caption{Diversity challenges.}
		\label{fig3c}
	\end{subfigure}
	
	\caption{\cref{fig3a}~$\sim$~\cref{fig3c} shows the pedestrian models, camera deployment, and challenges during the CARGO construction, respectively. In \cref{fig3b}, ``A-Cam'' and ``G-Cam'' represent the aerial and ground cameras, where the yellow sectors represent the view range of the ground cameras, and the green arrows represent the motion strategy of the aerial cameras. The challenges displayed in \cref{fig3c} are view variation, illumination variation, occlusion, and resolution variation from top to bottom.} 
	\label{fig3}
\end{figure*}

\subsection{Optimization}
To meet the task goal (\cref{goal}), the objective of $\mathcal{F}(\cdot, \theta_\mathcal{F})$ consists of three components: identity classifier, view classifier, and orthogonal loss. 

For each $x_i \in B$, the identity classifier utilizes both the cross-entropy  loss and the triplet loss to supervise $t_m^{N+1}(x_i)$, which can be written as:
\begin{equation}
	\mathcal{L}_{i}^{c}=\small{\frac{1}{|B|}}\sum_{i=1}^{|B|}{y_i\log \left( \hat{y}_i \right)},
\end{equation}
where $|B|$ represents the batch size, $y_i$ represents the identity label, and $\hat{y}_i$ denotes the identity prediction by $t_m^{N+1}(x_i)$.  The triplet loss is defined as follows:
\begin{equation}
	\mathcal{L}_{i}^{t}=\small{\frac{1}{|B|}}\sum_{i=1}^{|B|}{\left[ \begin{array}{c}
			\left\| t_{m}^{N+1}(x_i)-t_{m}^{N+1}(x_{i}^{+}) \right\| _{2}^{2}-\\
			\left\| t_{m}^{N+1}(x_i)-t_{m}^{N+1}(x_{i}^{-}) \right\| _{2}^{2}+\alpha\\
		\end{array} \right] _+},
\end{equation}
where $x_{i}^{+}/x_{i}^{-}$ denotes the hard positive/negative sample of $x_i$ in the current $B$,  $\left[\cdot\right]_+$ stands for $max(\cdot, 0)$,  $\alpha$ is the hyperparameter of the margin.

View classifier supervises $t_v^{N+1}(x_i)$ with only the cross-entropy loss, which can be written as:
\begin{equation}
	\mathcal{L}_{v}^{c}=\small{\frac{1}{|B|}}\sum_{i=1}^{|B|}{v_i\log \left( \hat{v}_i \right)},
\end{equation}
where $v_i$ represents the view label of $x_i$, and $\hat{v}_i$ denotes the view prediction by $t_v^{N+1}(x_i)$.

To enable $t_m^{N+1}(x_i)$ and $t_v^{N+1}(x_i)$ to be truly decoupled, we impose an orthogonality loss that makes them less similar to each other, which can be written as:
\begin{equation}
	\mathcal{L}_o=\small{\frac{1}{|B|}}\sum_{i=1}^{|B|}{\frac{\left| \left< t_{m}^{N+1}(x_i),t_{v}^{N+1}(x_i) \right> \right|}{\left\| t_{m}^{N+1}(x_i) \right\|_2 \cdot \left\| t_{v}^{N+1}(x_i) \right\|_2}},
	\label{oreg}
\end{equation}
where $\left| \left<\cdot, \cdot \right> \right|$ represents the absolute value after the dot product of two token embeddings.\footnote{The output of $\left|\cdot\right|$ is the count of elements when the input is a set, and the output is the absolute value when the input is a number.} We also provide a clear illustration of \cref{oreg} in \cref{fig:vdt} (the gray rectangle in the middle on the right). Please note that the dashed blue and the yellow vector represent the meta token and view token outputted by \cref{attn} in $\mathcal{F}_N$, respectively, and the blue vector represents the result of \cref{innersub}, which is the updated $t_{m}^{N+1}(x_i)$ outputted by $\mathcal{F}_N$. It is shown that achieving decoupling requires a two-part collaboration, where \cref{innersub} separates view-related features from global features, and \cref{oreg} constrains the remaining view-unrelated features to be orthogonally independent of view-related features. Overall, the total objective of VDT can be written as:
\begin{equation}
	\mathcal{L} = \mathcal{L}_{i}^{c} + \mathcal{L}_{i}^{t} + \lambda(\mathcal{L}_{v}^{c} + \mathcal{L}_{o}), \label{total_loss}
\end{equation}
where $\lambda$ is a hyperparameter to balance multiple objectives.

\begin{table*}[t]
	\centering
	\caption{Performance comparison of the mainstream methods under four settings of the proposed CARGO dataset. ``ALL'' denotes the overall retrieval performance of each method. ``G$\leftrightarrow$G'', ``A$\leftrightarrow$A'', and ``A$\leftrightarrow$G'' represent the performance of each model in several specific retrieval patterns. Rank1, mAP, and mINP are reported (\%). The best performance is shown in \textbf{bold}.}
	\label{performance-cargo}
	\renewcommand{\arraystretch}{1.2}
	\resizebox{\textwidth}{!}{%
		\begin{tabular}{|c|ccc|ccc|ccc|ccc|}
			\hline
			\multirow{2}{*}{Method} & \multicolumn{3}{c|}{Protocol 1: ALL} & \multicolumn{3}{c|}{Protocol 2: G$\leftrightarrow$G} & \multicolumn{3}{c|}{Protocol 3: A$\leftrightarrow$A} & \multicolumn{3}{c|}{Protocol 4: A$\leftrightarrow$G} \\ \cline{2-13} 
			& Rank1 & mAP & mINP & Rank1 & mAP & mINP & Rank1 & mAP & mINP & Rank1 & mAP & mINP \\ \hline
			SBS \cite{SBS} & 50.32 & 43.09 & 29.76 & 72.31 & 62.99 & 48.24 & 67.50 & 49.73 & 29.32 & 31.25 & 29.00 & 18.71 \\
			PCB \cite{pcb_TPAMI} & 51.00 & 44.50 & 32.20 & 74.10 & 67.60 & 55.10 & 55.00 & 44.60 & 27.00 & 34.40 & 30.40 & 20.10 \\
			BoT \cite{BoT} & 54.81 & 46.49 & 32.40 & 77.68 & 66.47 & 51.34 & 65.00 & 49.79 & 29.82 & 36.25 & 32.56 & 21.46 \\
			MGN \cite{MGN} & 54.81 & 49.08 & 36.52 & \textbf{83.93} & 71.05 & 55.20 & 65.00 & 52.96 & 36.78 & 31.87 & 33.47 & 24.64 \\
			VV \cite{r1_1, r1_2} & 45.83 & 38.84 & 39.57 & 72.31 & 62.99 & 48.24 & 67.50 & 49.73 & 29.32 & 31.25 & 29.00 & 18.71 \\
			AGW \cite{AGW} & 60.26 & 53.44 & 40.22 & 81.25 & \textbf{71.66} & 58.09 & 67.50 & 56.48 & 40.40 & 43.57 & 40.90 & 29.39 \\ \hline
			ViT \cite{Vit} & 61.54 & 53.54 & 39.62 & 82.14 & 71.34 & 57.55 & 80.00 & 64.47 & 47.07 & 43.13 & 40.11 & 28.20 \\
			\textbf{VDT (Ours)} & \textbf{64.10} & \textbf{55.20} & \textbf{41.13} & 82.14 & 71.59 & \textbf{58.39} & \textbf{82.50} & \textbf{66.83} & \textbf{50.22} & \textbf{48.12} & \textbf{42.76} & \textbf{29.95} \\ \hline
		\end{tabular}%
	}
\end{table*}

\begin{table}[t]
	\centering
	\caption{Performance comparison of the mainstream methods under two settings of AG-ReID dataset. ``A$\leftrightarrow$G'', and ``G$\leftrightarrow$A'' represent the performance in two specific patterns. Rank1, mAP, and mINP are reported (\%). The best performance is shown in \textbf{bold}.}
	\label{performance-agreid}
	\renewcommand{\arraystretch}{1.2}
	\resizebox{0.47\textwidth}{!}{%
		\begin{tabular}{|c|ccc|ccc|}
			\hline
			\multirow{2}{*}{Method} & \multicolumn{3}{c|}{Protocol 1: A$\rightarrow$G} & \multicolumn{3}{c|}{Protocol 2: G$\rightarrow$A} \\ \cline{2-7} 
			& Rank1 & mAP & mINP & Rank1 & mAP & mINP \\ \hline
			SBS \cite{SBS} & 73.54 & 59.77 & - & 73.70 & 62.27 & - \\
			BoT \cite{BoT} & 70.01 & 55.47 & - & 71.20 & 58.83 & - \\
			OSNet \cite{OSNet-AIN} & 72.59 & 58.32 & - & 74.22 & 60.99 & - \\
			VV \cite{r1_1, r1_2} & 77.22 & 67.23 & 41.43 & 79.73 & 69.83 & 42.37 \\
			\hline
			ViT \cite{Vit} & 81.28 & 72.38 & - & 82.64 & 73.35 & - \\
			Explain \cite{AG-ReID} & 81.47 & 72.61 & - & 82.85 & 73.39 & - \\ 
			\textbf{VDT (Ours)} & \textbf{82.91} & \textbf{74.44} & \textbf{51.06} &\textbf{86.59} &\textbf{78.57} & \textbf{52.87} \\ \hline
		\end{tabular}%
	}
\end{table}

\section{Dataset: CARGO}
\subsection{Motivation}
\cref{datasets} compares the statistics and several other details of various ReID datasets, demonstrating the gap between the homogeneous and heterogeneous views. First, the dataset scale of the homogeneous view has far exceeded that of the heterogeneous view. For example, the identities in the latest AG-ReID \cite{AG-ReID} are about a quarter of that of the early Market1501 \cite{market}. Second, synthetic data has been well-studied in the homogeneous view, and benefiting from low construction costs, synthetic datasets tend to contain more identities and images. However, the synthetic data study is absent in the view-heterogeneous ReID. The above limitations motivate us to contribute a large-scale synthetic dataset in the heterogeneous view. As can be seen from \cref{datasets}, the identities and images of CARGO are 12.8 and 4.9 times larger than that of AG-ReID, which brings the view-heterogeneous ReID to a comparable scale with the view-homogeneous ReID.

\subsection{Description}
The construction of CARGO is summarized as the construction of human models, the deployment of the camera network, and the data collection, shown in \cref{fig3}.

First, MakeHuman \cite{makehuman} is adopted to create person models with different identities by assigning randomly different values to predefined attributes, such as gender, height, age, body shape,  skin color, hair, clothing, \etc. We have created 5,000 pedestrians for subsequent operations, and some models have been shown in \cref{fig3a}.

Second, Unity3D \cite{unity} is utilized to simulate a real-world streetscape. As shown in \cref{fig3b}, we pick three blocks to deploy the camera network, where a ground camera network (``G-Cam'' area) serves relatively convenient areas such as downtown, and two aerial camera networks (``A-Cam'' area) serve the suburbs of a city. In the ``G-Cam'' area, we deploy eight fixed cameras at each street corner. In the ``A-Cam'' area, we design different drone roaming strategies based on the size of the surveillance area. For the small area (left area), we deploy one drone with a $90^\circ$ overhead view, allowing it to move counterclockwise around each street. For a large area (right area), we deploy individual drones on each of the four streets with a $45^\circ\sim60^\circ$ tilt view, allowing them to move back and forth on corresponding streets. 

Finally, we import human models into the streetscape and further move them under the ground and aerial networks. We allow multiple pedestrians to move simultaneously to simulate the random occlusion among pedestrians in the real world. Meanwhile, we collect valid images of the target persons if they appear in cameras without severe occlusion (train $\leq 70\%$, and test $\leq 75\%$). In total, CARGO contains 108,563 person images from 13 cameras. 

\subsection{Challenge}
\cref{fig3c} illustrates abundant challenges prevalent in real-world scenarios in the CARGO dataset. The most significant challenge is the view discrepancy from different types of cameras, which produces a vast variation and rotation of person postures. Besides, \cref{fig3c} and \cref{datasets} show another change in the camera height, from 5 meters for the lowest ground camera to 75 meters for the highest $90^\circ$ view drone, resulting in a huge resolution variation. The maximum resolution in CARGO is $1009\times539$, while the minimum is only $22\times7$. Moreover, CARGO also contains other complex illumination and occlusion variations, \etc. Overall, CARGO simulates and covers major real-world scenarios and challenges and can be considered a large-scale benchmark.


\begin{figure*}[t]
	\centering
	\begin{subfigure}{0.33\linewidth}
		\centering
		\includegraphics[width=\linewidth]{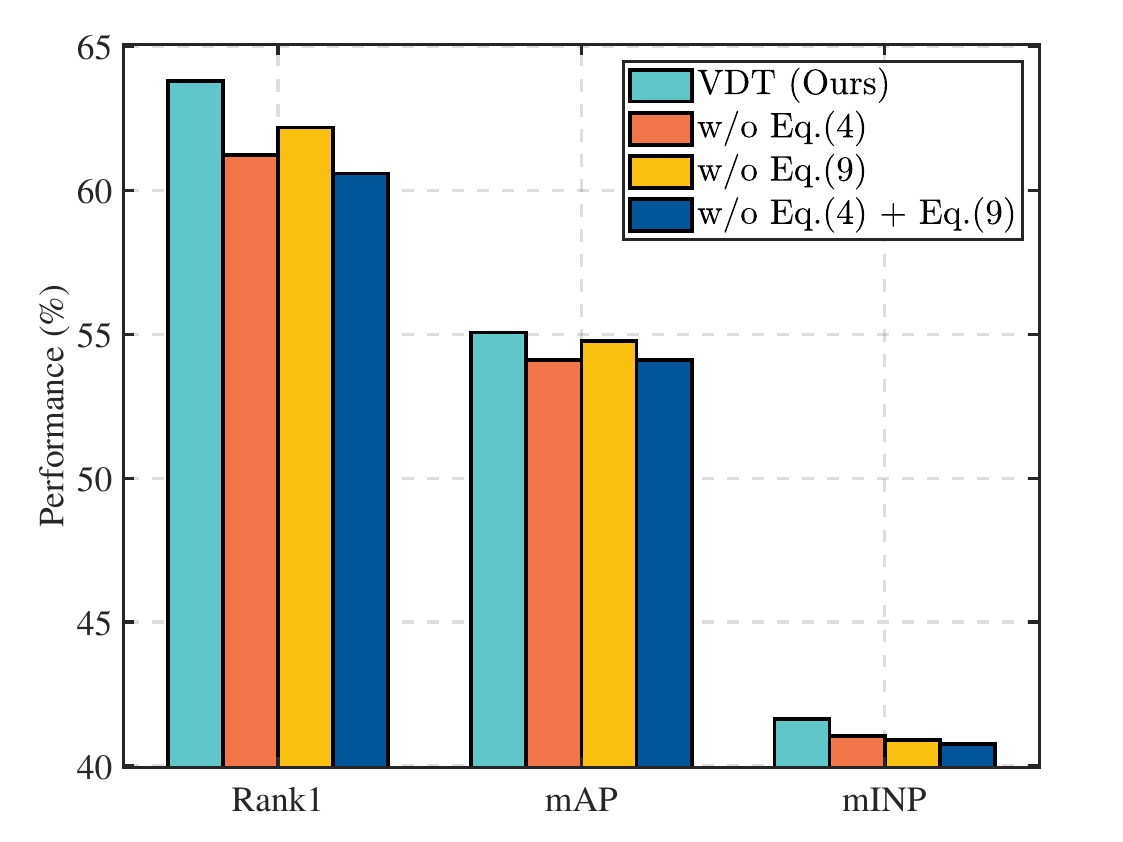}
		\caption{CARGO under Protocol 1.}
		\label{ab1}
	\end{subfigure} 
	\begin{subfigure}{0.33\linewidth}
		\centering
		\includegraphics[width=\linewidth]{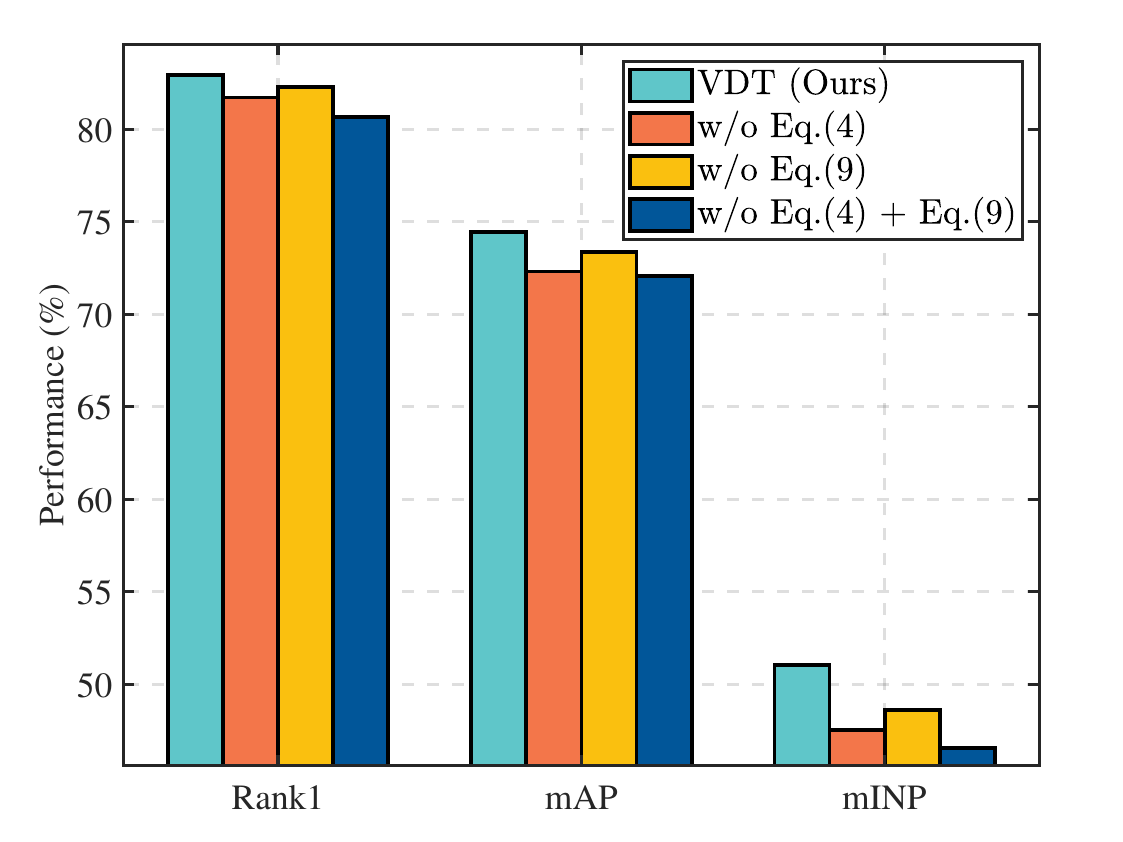}
		\caption{AG-ReID under Protocol 1.}
		\label{ab2}
	\end{subfigure} 
	\begin{subfigure}{0.33\linewidth}
		\centering
		\includegraphics[width=\linewidth]{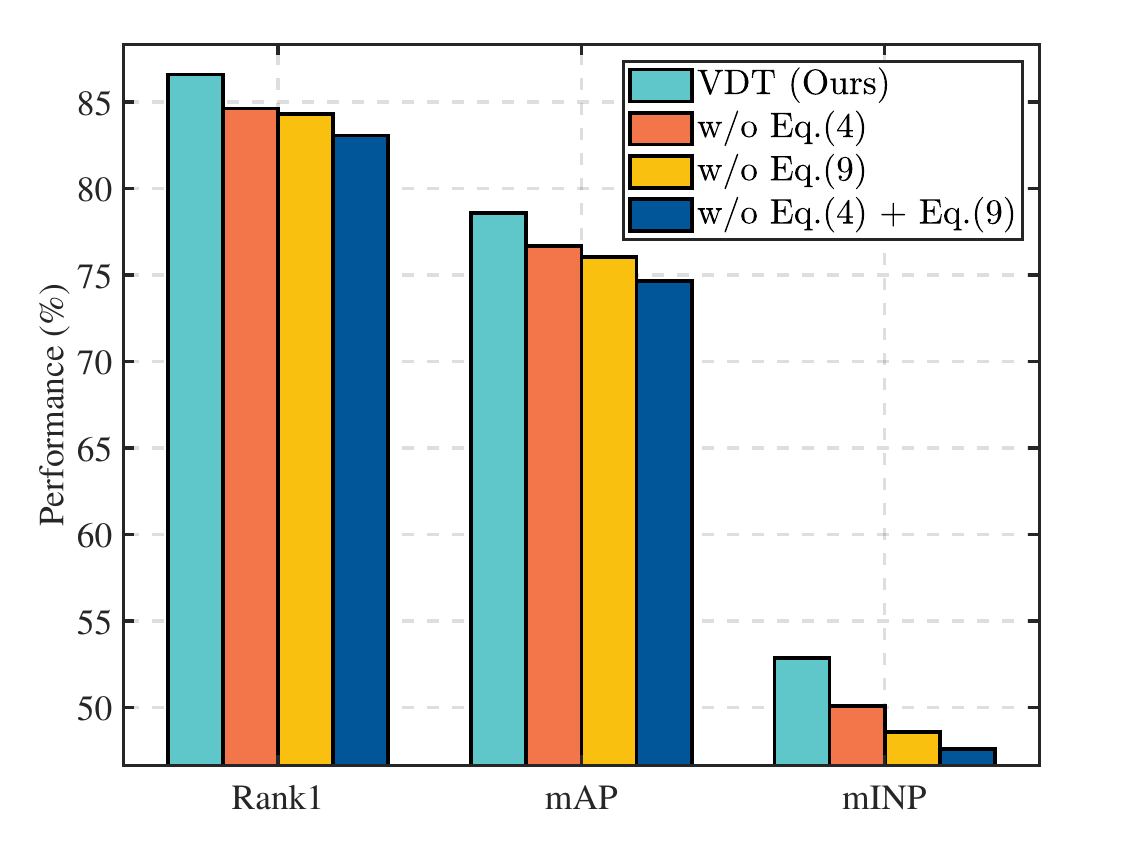}
		\caption{AG-ReID under Protocol 2.}
		\label{ab3}
	\end{subfigure}
	
	\caption{\cref{ab1}~$\sim$~\cref{ab3} show the ablation experiments about orthogonal decoupling of view-related and view-unrelated features in the VDT on two dataset, which consists of two important parts, \cref{innersub} and \cref{oreg}. Rank1, mAP, and mINP are reported (\%).} 
	\label{ab}
\end{figure*}

\section{Experiment}
Importantly, please refer to the \textit{supplementary material} for detailed experiment settings, visualization of retrieval and feature, cross-dataset evaluation, and other discussions.

\subsection{Dataset and Metric}
\textbf{Dataset.} Two datasets are adopted for evaluation, CARGO and AG-ReID \cite{AG-ReID}. Some statistics regarding these datasets are shown in \cref{datasets}.  As for CARGO, we split it into the train (51,451 images with 2500 IDs) and test sets (51,024 images with the remaining 2500 IDs) with an almost 1:1 ratio. In the training set, each person has an average of 10 images and on average moves across 6.47 ground and 3.61 aerial cameras. In the test set, we select 149 IDs as the query set, and all the remaining IDs are placed in the gallery as distractors. For each identity selected as a query, we keep only its images from two random cameras and use the image from one camera as the query and the image from the other as the gallery. On this basis, we design four protocols (ALL, A$\leftrightarrow$A, G$\leftrightarrow$G, and A$\leftrightarrow$G) to adequately evaluate the model performances, where ``ALL'' focuses on the comprehensive retrieval performance, and the latter three focus on some specific retrieval patterns. A$\leftrightarrow$A only retains the data under the aerial camera in the test set for evaluation (60 query IDs with 134 images, 2404 gallery IDs with 18,444 images), same for G$\leftrightarrow$G (89 query IDs with 178 images, 2447 gallery IDs with 32,268 images). A$\leftrightarrow$G relabels the original test set into two domains (aerial and ground domain) based on the view label. The training set of all testing protocols retains same.

As for AG-ReID, there are 11,554 images with 199 IDs for training and 12,464 images with 189 IDs for testing following \cite{AG-ReID}. The test set contains 2,033 query images and 10,429 gallery images. There are two protocols in AG-ReID, denoted as A$\rightarrow$G and G$\rightarrow$A, where the former contains 1,701 aerial query images and 3,331 ground gallery images for 189 identities, and the latter contains 962 ground query images and 7,204 aerial gallery images for 189 identities. Optionally, AG-ReID also provides detailed person attribute labeling as additional prior information.

\noindent\textbf{Metric.} The cumulative matching characteristic at Rank1,  mean Average Precision (mAP), and mean Inverse
Negative Penalty (mINP) \cite{AGW} are adopted as the evaluation metrics. 

\begin{figure*}[t]
	\centering
	\begin{subfigure}{0.33\linewidth}
		\centering
		\includegraphics[width=\linewidth]{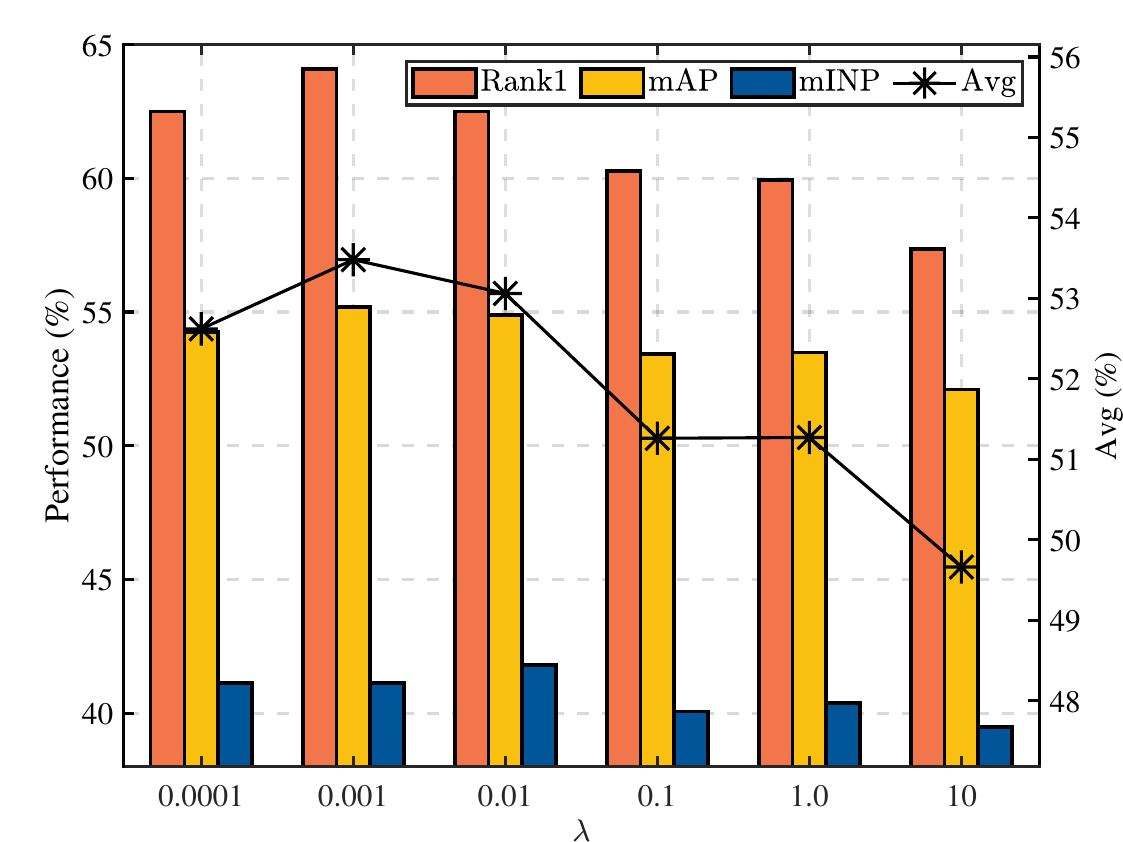}
		\caption{CARGO under Protocol 1.}
		\label{para1}
	\end{subfigure} 
	\begin{subfigure}{0.33\linewidth}
		\centering
		\includegraphics[width=\linewidth]{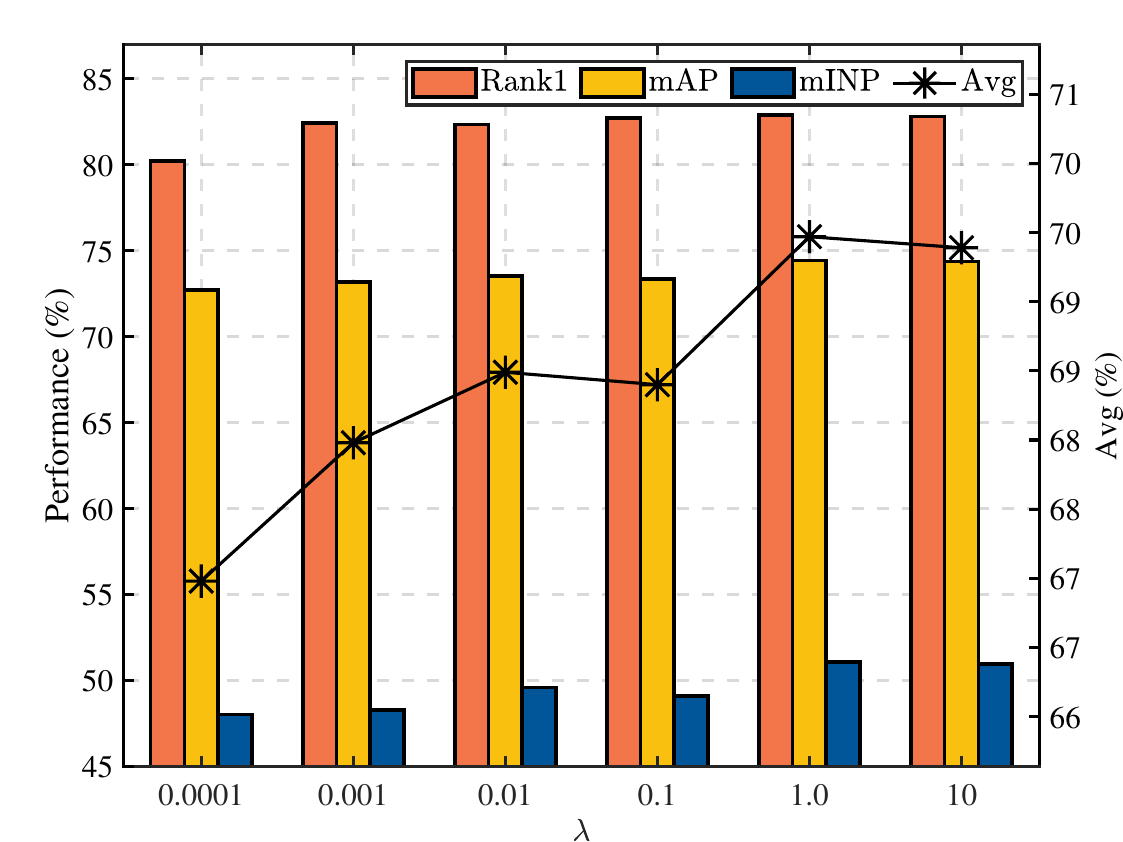}
		\caption{AG-ReID under Protocol 1.}
		\label{para2}
	\end{subfigure} 
	\begin{subfigure}{0.33\linewidth}
		\centering
		\includegraphics[width=\linewidth]{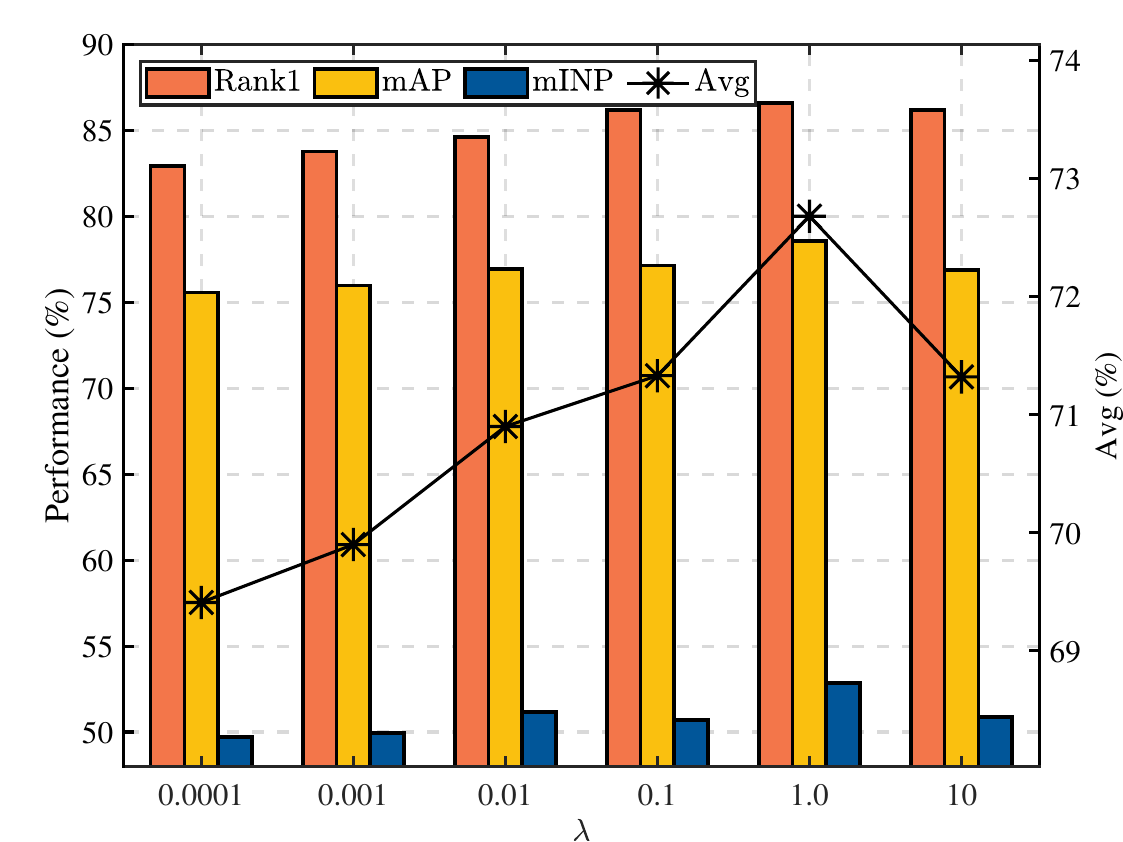}
		\caption{AG-ReID under Protocol 2.}
		\label{para3}
	\end{subfigure}
	
	\caption{\cref{para1}~$\sim$~\cref{para3} show the relationship between $\lambda$ and performance on two datasets. For simplicity, only protocol 1 is shown on the CARGO dataset. Rank1, mAP, and mINP are reported (\%). Avg represents the average performance of Rank1, mAP, and mINP.} 
	\label{para}
\end{figure*}

\subsection{Performance}
\noindent\textbf{Accuracy.} We compare the proposed VDT with other competitive methods on CARGO in \cref{performance-cargo}, including CNN-based (BoT \cite{BoT}, SBS \cite{SBS}, VV \cite{r1_1, r1_2}, MGN \cite{MGN}, AGW \cite{AGW}) and transformer-based methods (ViT \cite{Vit}), where VV means a  vehicle ReID baseline with vanilla batch-based triplet loss and diverse viewpoint batch selection. Similar experimental results on AG-ReID are reported in \cref{performance-agreid}, where  Explain~\cite{AG-ReID} relies on pedestrian attribute priors and therefore does not generalize well to CARGO. Several performances in \cref{performance-agreid} follow the results reported in \cite{AG-ReID}.  We can draw the following conclusions:  (1) The proposed VDT achieves the state-of-the-art performances that are clearly improved over the baseline, especially in heterogeneous matching. For example, VDT surpasses the mAP/Rank1/mINP baseline by 4.99\%/2.65\%/1.75\% on the A$\leftrightarrow$G of CARGO. Besides, VDT also brings different degree of benefits to other CARGO  protocols. VDT also surpasses Explain on mAP/Rank1 by 3.74\%/5.18\% on the G$\rightarrow$A of AG-ReID, which strongly demonstrates the importance of view decoupled operations in AGPReID to mitigate the disruption of identity representation by view bias whether in homogeneous or heterogeneous matching. (2) Previous view-homogeneous methods show varying degrees of performance degradation in AGPReID, especially on view-heterogeneous protocols, indicating that view bias causes the identity feature under different views to behave with poor cohesion. However, the above methods ignore this issue. (3) The superior performance of VDT does not come from a strong baseline but from the proposed method itself. On the CARGO dataset, the baseline does not consistently outperform the CNN-based AGW on all metrics; on the AGPReID dataset, the baseline is also weaker than the previous Explain. However, VDT achieves the SOTA performance on two datasets, especially on AG-ReID, using fewer prior labels and achieving better performance. 

\noindent\textbf{Speed.} We report the inference time for baseline and VDT for retrieving a single image, which is 2.27ms and 2.28ms on the CARGO dataset and 1.44ms and 1.46ms on the AG-ReID dataset, respectively. These results validate the previous complexity analysis in \cref{sec3.2} and show that VDT does not add additional time consumption. Slight time fluctuations tend to be acceptable, due to hardware devices.

\subsection{Ablation Study}
As described in \cref{sec3.2}, the core contribution of VDT lies in the decoupling of identity and view features by \cref{innersub} and \cref{oreg}. Therefore, we aim to demonstrate the contribution of each part to VDT, which has been shown in \cref{ab}. Similar trends in the two datasets demonstrate that the absence of each part leads to a degradation of the model's performance, and the absence of both simultaneously leads to a more severe performance degradation. 

By comparing Row1 and Row2, the absence of \cref{innersub} leads to the degradation of $t_m$ and $t_v$ to be equivalent, focusing on the identity and view features in the image respectively, and constraining them to be independent through \cref{oreg}. However, lacking the guidance of explicit separation within VDT, the decoupling goal cannot be effectively achieved through \cref{oreg} alone. Similarly, by comparing Row1 and Row3, the identity and view features outputted by  VDT are not guaranteed to be orthogonally separated from each other if \cref{oreg} is missing, and thus  $t_m$ containing a minor view bias is still not strongly discriminative.

\subsection{Parameter Analysis}
As described in \cref{sec3.2}, VDT  introduces only one hyperparameter $\lambda$ in \cref{total_loss} that needs to be manually tuned for multi-task balance. We select six magnitudes of $\lambda$ from 1$e$-4 to 10 and  the corresponding results are shown in \cref{para}. To show a clear tendency, we show the average performance of Rank1, mAP, and mINP at each $\lambda$. Although the performance shows the same trend of increasing and then decreasing on both datasets as $\lambda$ increases, the peak performance is achieved when $\lambda=0.001$ on CARGO (\cref{para1}) and $\lambda=1.0$ on AG-ReID (\cref{para2} and \cref{para3}). 

The key of this phenomenon is the synchronous updating between identity and view classifier, because the view classifier is a simple binary classification, but the identity classifier on the two datasets shows different difficulties depending on the number of identities. As shown in \cref{datasets}, CARGO owns a thousand-level identities, so the identity classification is relatively complicated, and in order to keep tasks in \cref{total_loss} able to update synchronously, a small $\lambda$ is more appropriate for view-related objectives. Similarly, a large $\lambda$ is selected because of a hundred-level identities in AGPReID. We also provide an empirical $\lambda$ selection guideline to avoid exhaustive search based on above analysis: $\lambda$ tends to be smaller for datasets with larger identity scale.
\section{Conclusion and Future Work}
This paper focuses on ReID under a view-heterogeneous scenario, aerial-ground ReID (AGPReID). First, the view-decoupled transformer (VDT) is proposed specifically for the dramatic view differences in AGPReID, which decouples view-related and view-unrelated features by two major operations: hierarchically subtractive separation and orthogonal loss. Second, we contribute a large-scale dataset called CARGO, in which the identities and images are 12.8 and 4.9 times larger than the previous dataset. Experiments on two datasets demonstrate the superiority of VDT and the necessity of view decoupling in AGPReID.

Although view discrepancy is the most significant factor in AGPReID, it is not the only source of disruptions. Therefore,  further exploration of discriminative identity representations under multiple disturbances in the aerial-ground camera network is a promising avenue of future research.

\section*{Acknowledgments} This project was supported in part by the NSFC (U22A2095, 62076258), in part by the Key-Area Research and Development Program of Guangzhou (202206030003), in part by Guangdong Project (No. 2020B1515120085), and in part by International Program Fund for Young Talent Scientific Research People, Sun Yat-Sen University.

{
	\clearpage
    \small
    \bibliographystyle{unsrt}
    \bibliography{main}
}

\clearpage
\setcounter{page}{1}
\maketitlesupplementary

\section{Experiments}

\subsection{Setting}
VDT adopts the ViT-base \cite{Vit} as the baseline, which contains $N=12$ encoder blocks and is pre-trained on the ImageNet \cite{ImageNet}. The patch size and stride size in VDT are set to 16$\times$16. The size of the input image is resized to 256$\times$128, so the $M=128$. The embedding shape $d$ of tokens is set to 768. The $t_m$ and $t_v$ are randomly initialized at the beginning of training. During training, we adopt padding with 10 pixels, random cropping, and random erasing with a probability of 0.5 for data augmentation. We adopt a soft version of triplet \cite{AGW} to avoid manually selecting $m$ in the triplet loss. The stochastic gradient descent \cite{sgd} optimizer is used. The cosine learning rate decay is adopted to reduce the learning rate from initial $8\times10^{-3}$ to final $1.6\times10^{-6}$. The number of training epochs is 120. The batch size is 128, including 32 identities, each with four images. We do not apply any data augmentation or re-ranking during inference. The VDT is implemented by PyTorch \cite{pytorch, SBS}. All experiments have been conducted on one A5000 GPU. 

\subsection{Visualization}
\begin{figure}[t]
	\centering
	\includegraphics[width=\linewidth]{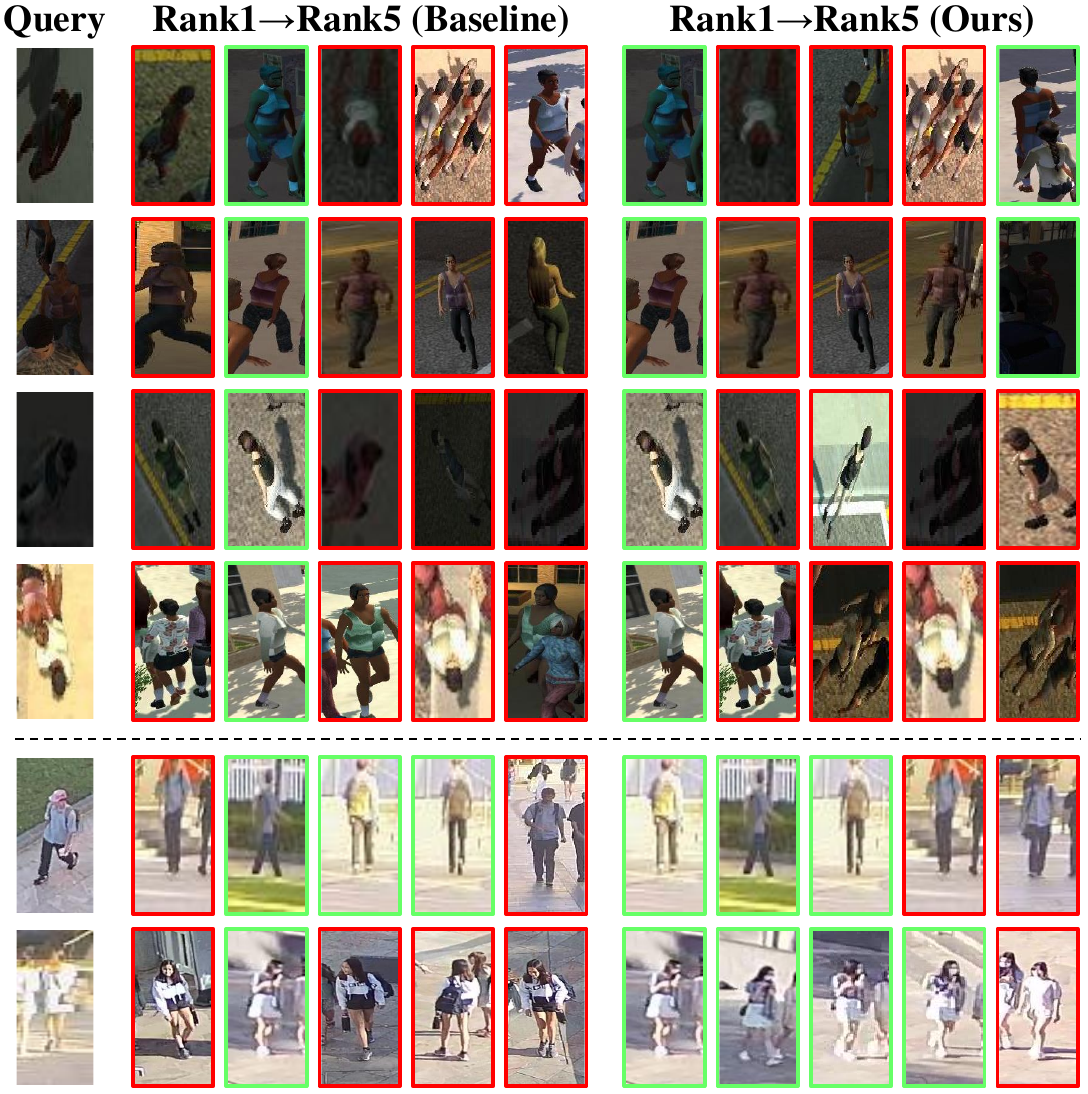}
	\caption{Comparison of several retrieval visualizations on the CARGO and AG-ReID dataset protocols. Red and green boxes represent wrong and correct matchings. The top five are listed.}
	\label{vis}
\end{figure}
\noindent\textbf{Retrieval visualization.} \cref{vis} shows the retrieval advantages of VDT under multiple protocols of the two datasets. Compared to the baseline, VDT achieves better feature decoupling and extracts more discriminative descriptions of the target person from view-unrelated features, which makes the identity features more robust under each protocol of both datasets. \cref{vis} amply illustrates that the proposed view decoupling is feasible and effective for alleviating view discrepancy in AGPReID.
\begin{figure}[t]
	\centering
	\begin{subfigure}[b]{0.2\textwidth}
		\includegraphics[width=\textwidth]{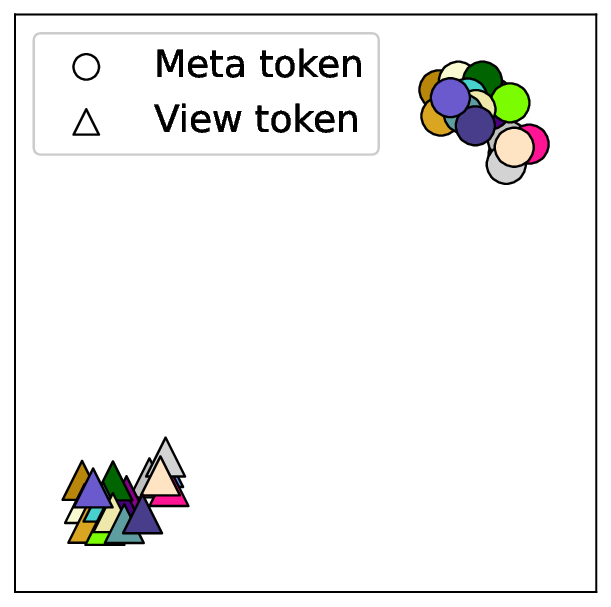}
		\caption{CARGO dataset}
		\label{fig:sub1}
	\end{subfigure}
	\hfill 
	\begin{subfigure}[b]{0.2\textwidth}
		\includegraphics[width=\textwidth]{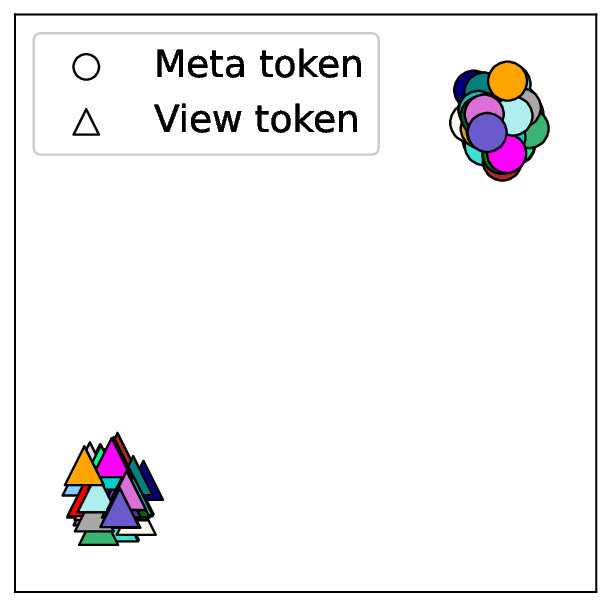}
		\caption{AG-ReID dataset}
		\label{fig:sub2}
	\end{subfigure}
	\caption{Visualization of meta and view tokens via tSNE.}
	\label{e3}
\end{figure}

\noindent\textbf{Feature visualization.} As shown in \cref{e3}, we randomly select a pedestrian identity on each dataset and visualize the meta (circle) and view (triangles) tokens corresponding to all the images under this identity (the same color means from the same image). In \cref{e3}, meta and view tokens show good cohesion and significant differences from each other, which indicates that our method achieves good decoupled representations between these two tokens.

\subsection{Cross-dataset evaluation} 
\begin{table}[t]
	\centering
	\caption{Cross-domain performance evaluations (\%) for transferring from CARGO to AG-ReID dataset.}
	\label{e1}
	\renewcommand{\arraystretch}{1.2}
	\resizebox{0.45\textwidth}{!}{%
		\begin{tabular}{|c|cccccc|}
			\hline
			\multirow{3}{*}{Method} & \multicolumn{6}{c|}{CARGO→AG-ReID} \\ \cline{2-7} 
			& \multicolumn{3}{c|}{Protocol1:   A→G} & \multicolumn{3}{c|}{Protocol2: G→A} \\ \cline{2-7} 
			& Rank1 & mAP & \multicolumn{1}{c|}{mINP} & Rank1 & mAP & mINP \\ \hline
			ViT \cite{Vit} & 1.59 & 1.95 & \multicolumn{1}{c|}{0.8} & 3.01 & 2.31 & 0.95 \\ \hline
			\textbf{VDT (Ours)} & \textbf{19.33} & \textbf{11.81} & \multicolumn{1}{c|}{\textbf{1.63}} & \textbf{15.38} & \textbf{11.73} & \textbf{3.38} \\ \hline
		\end{tabular}%
	}
\end{table}
The results (training on CARGO, and testing on AR-ReID) have been shown in \cref{e1}, which indicates that the direct cross-dataset (domain) evaluation is challenging, but our VDT has advantages over the baseline. One possible reason is that our decoupling strategy allows identity-related learning to be less perturbed by domain-related factors (e.g., view bias), leading to more discriminative identity features.

\begin{table*}[t]
\centering
\caption{Performance comparison (\%) on CARGO and AG-ReID datasets.}
\label{e2}
\resizebox{0.9\textwidth}{!}{%
\begin{tabular}{|c|ccc|ccc|ccc|}
\hline
\multirow{3}{*}{Method} & \multicolumn{3}{c|}{CARGO} & \multicolumn{3}{c|}{AG-ReID} & \multicolumn{3}{c|}{AG-ReID} \\ \cline{2-10} 
 & \multicolumn{3}{c|}{Protocol1: ALL} & \multicolumn{3}{c|}{Protocol1: A→G} & \multicolumn{3}{c|}{Protocol2: G→A} \\ \cline{2-10} 
 & Rank1 & mAP & mINP & Rank1 & mAP & mINP & Rank1 & mAP & mINP \\ \hline
TransReID \cite{TransReID} & 60.90 & 53.17 & 39.57 & 78.25 & 70.03 & 44.74 & 79.74 & 70.79 & 45.12 \\ \hline
\textbf{VDT (Ours)} & \textbf{64.10} & \textbf{55.20} & \textbf{41.13} & \textbf{82.91} & \textbf{74.44} & \textbf{51.06} & \textbf{86.59} & \textbf{78.57} & \textbf{52.87} \\ \hline
\end{tabular}%
}
\end{table*}

\subsection{Discussion}
The results of TransReID \cite{TransReID} on two datasets have been shown in \cref{e2}, which shows that show that although TransReID \cite{TransReID} introduces additional camera information and encodes it as part of the input, its performance on both datasets is still weaker than that of the our method, proving the effectiveness of the proposed decoupling strategy proposed for the AGPReID task.

\end{document}